\pdfoutput=1

\documentclass[11pt]{article}
\usepackage{authblk}
\usepackage[dvipsnames]{xcolor}

\usepackage{acl}

\usepackage{times}
\usepackage{latexsym}

\usepackage[T1]{fontenc}

\usepackage[utf8]{inputenc}
\usepackage{microtype}
\usepackage{url}
\usepackage{times}
\usepackage{latexsym}
\usepackage{tabularx}
\usepackage{graphicx}
\usepackage{booktabs}
\usepackage{makecell}
\usepackage{amsfonts}       %
\usepackage{nicefrac}       %
\usepackage{microtype}      %
\usepackage{multirow}
\usepackage{caption}
\usepackage{amsmath}
\usepackage{float} 
\usepackage{ragged2e}

\usepackage{enumitem}
\mathchardef\mhyphen="2D

\usepackage{microtype}

\usepackage{soul}
\newcommand{\hlc}[2][yellow]{{\sethlcolor{#1}\hl{#2}}}

\newif\ifcomments
\commentstrue

\ifcomments
    \newcommand*{\TODO}[1]{\textcolor{red}{[TODO: #1]}}
    \newcommand*{\tocite}[1]{(\textcolor{blue}{#1})}
    \newcommand*{\tocitep}[1]{(\textcolor{blue}{#1})}
    \newcommand*{\tocitet}[1]{\textcolor{blue}{#1}}
    
    \newcommand*{\bhargavi}[1]{\textcolor{orange}{[Bhargavi: #1]}}
    \newcommand*{\ian}[1]{\textcolor{olive}{[Ian: #1]}}
    \newcommand*{\matt}[1]{\textcolor{purple}{[Marr: #1]}}

    \newcommand*{\maybedelete}[1]{\textcolor{red}{\sout{#1}}}
\else
    \newcommand*{\TODO}[1]{}
    \newcommand*{\tocite}[1]{}
    \newcommand*{\tocitep}[1]{}
    \newcommand*{\tocitet}[1]{}
    
    \newcommand*{\bhargavi}[1]{}
    \newcommand*{\ian}[1]{}
    \newcommand*{\matt}[1]{}
    \newcommand*{\iansidenote}[1]{}
    
    \newcommand*{\maybedelete}[1]{}
    
\fi

\renewcommand\footnotemark{}
\makeatletter
\renewcommand\AB@authnote[1]{\rlap{\textsuperscript{\normalfont#1}}}

\makeatother

\title{Retrieval-guided Counterfactual Generation for QA}

\author[1$*$]{Bhargavi Paranjape\thanks{$^*$Work performed during an internship at Google.}}
\author[2]{Matthew Lamm}
\author[2]{Ian Tenney}

\affil[1]{Paul G. Allen School of Computer Science \& Engineering, University of Washington}
\affil[2]{Google Research}

\affil[ ]{\texttt{bparan@cs.washington.edu}}
\affil[ ]{\texttt{\{mrlamm,iftenney\}@google.com}}

\begin{document}
\maketitle
\begin{abstract}
Deep NLP models have been shown to be brittle to input perturbations. Recent work has shown that data augmentation using counterfactuals --- i.e. minimally perturbed inputs --- can help ameliorate this weakness.
We focus on the task of creating counterfactuals for question answering, which presents unique challenges related to world knowledge, semantic diversity, and answerability.
To address these challenges, we develop a \emph{\textbf{R}etrieve-\textbf{G}enerate-\textbf{F}ilter} (RGF) technique to create counterfactual evaluation and training data with minimal human supervision. 
Using an open-domain QA framework and question generation model trained on original task data, we create counterfactuals that are fluent, semantically diverse, and automatically labeled.
Data augmentation with RGF counterfactuals improves performance on out-of-domain and challenging evaluation sets over and above existing methods, in both the reading comprehension and open-domain QA settings. Moreover, we find that RGF data leads to significant improvements to robustness to 
local perturbations.\footnote{Code at \url{https://github.com/google-research/language/tree/master/language/qa_counterfactuals}}
\end{abstract}

\section{Introduction}

Models for natural language understanding (NLU) may outperform humans on standard benchmarks, yet still often perform poorly under a multitude of distributional shifts (\citet{jia2017adversarial, naik2018stress, mccoy2019right}, \emph{inter alia}) due to over-reliance on spurious correlations or dataset artifacts. This behavior can be probed using counterfactual data \cite{kaushik2019learning, gardner-etal-2020-evaluating} designed to simulate interventions on specific attributes: for example, perturbing the movie review \textit{``A real stinker, one out of ten!"} to \textit{``A real classic, ten out of ten!"} allows us to discern the effect of adjective polarity on the model's prediction. Many recent works \citep[][\emph{inter alia}]{kaushik2019learning, kaushik-etal-2021-efficacy, wu-etal-2021-polyjuice, geva2021break} have shown that training augmented with this counterfactual data (CDA) improves out-of-domain generalization and robustness against spurious correlations. Consequently, several techniques have been proposed for the automatic generation of counterfactual data for several downstream tasks \cite{wu-etal-2021-polyjuice, ross2021tailor, ross-etal-2021-explaining, bitton2021automatic, geva2021break, asai2020logic, mille-etal-2021-another}.

\begin{figure}
    \centering
    \includegraphics[scale=0.50]{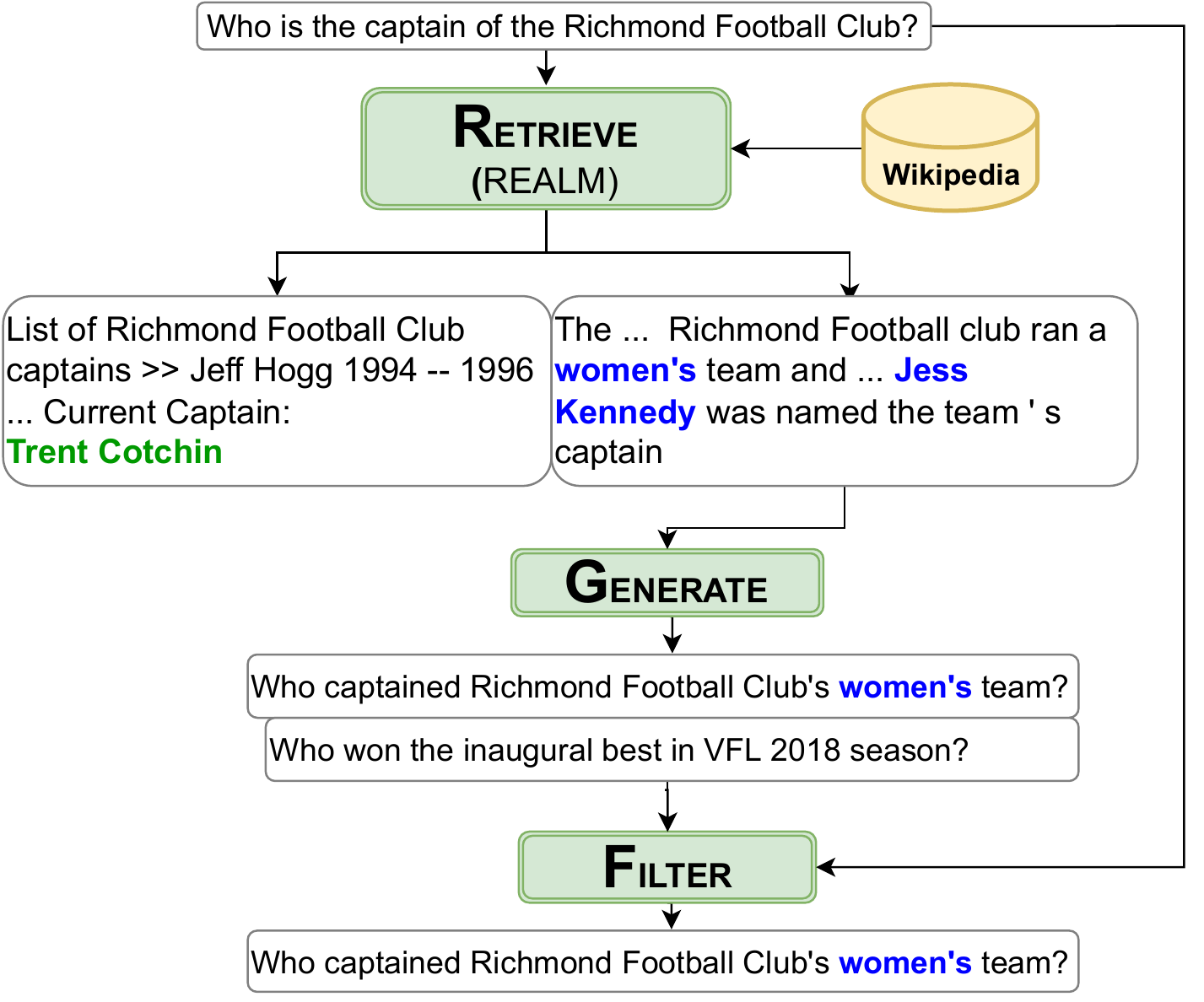}
    \caption{\emph{Retrieve-Generate-Filter} to generate counterfactual queries for Natural Question \cite{kwiatkowski2019natural} using an open-domain retrieval system, question generation and post-hoc filtering.}
    \label{fig:teaser}
\end{figure}

In this paper, we focus on counterfactual data for question answering, in both the reading comprehension and open-domain settings \citep[e.g.][]{rajpurkar2016squad,kwiatkowski2019natural}.
Model inputs consist of a question and optionally a context passage, and the target $a$ is a short answer span. 
Counterfactuals are often considered in the context of a specific causal model \citep{miller2019explanation,halpern2005causes}, but in this work we follow \citet{wu-etal-2021-polyjuice} and \citet{kaushik2019learning} and seek a method to generate counterfactuals that may be useful in many different settings. 
In QA, the set of possible causal features is large and difficult to specify \textit{a priori}; relevant factors are often instance-specific and exploring them may require world knowledge.
For example, going from \textit{``Who is the captain of the Richmond Football Club''} to a perturbed question \textit{``Who captained Richmond's women's team?''} as in Figure~\ref{fig:teaser} 
requires knowledge about the club's alternate teams, and the perturbation \textit{``Who was the captain of RFC in 1998?''} requires knowledge about the time-sensitive nature of the original question. In the absence of such knowledge, otherwise reasonable edits --- such as \textit{``Who captained the club in 2050?''} --- can result in false premises or unanswerable questions.

We develop a simple yet effective technique to address these challenges: \emph{\textbf{R}etrieve, \textbf{G}enerate, and \textbf{F}ilter} (RGF; Figure~\ref{fig:teaser}). We use the near-misses of a retrieve-and-read QA model to propose alternate contexts and answers which are closely related to --- but semantically distinct from --- the original question.
We then use a sequence-to-sequence question generation model \cite{alberti2019synthetic} to generate corresponding questions to these passages and answers.  
This results in fully-labeled examples, which can be used directly to augment training data or filtered post-hoc for analysis.

While our method requires no supervised inputs besides the original task training data, it is able to generate highly diverse counterfactuals covering a range of semantic phenomena (\S\ref{sec:intrinsic}), including many transformation types which existing methods generate through
heuristics \cite{dua2021learning}, meaning representations \citep{ross2021tailor,geva2021break} or human generation \citep{bartolo-etal-2020-beat, gardner-etal-2020-evaluating}. Compared to alternative sources of synthetic data (\S\ref{sec:baselines}), training augmented with RGF data improves performance on a variety of settings (\S\ref{results:rc}, \S\ref{results:odqa}), including out-of-domain \cite{fisch2019mrqa} and contrast evaluation sets \cite{bartolo-etal-2020-beat, gardner-etal-2020-evaluating}, while maintaining in-domain accuracy. Additionally, we introduce a measure of \textit{pairwise consistency}, and show that RGF significantly improves robustness to a range of local perturbations (\S\ref{sec:analysis}).

\section{Related Work}

\subsection{Counterfactual Generation}
There has been considerable interest in developing challenge sets for NLU that evaluate models on a wide variety of counterfactual scenarios.
\citet{gardner-etal-2020-evaluating, khashabi-etal-2020-bang, kaushik2019learning, ribeiro2020beyond} use humans to create these perturbations, optionally in an adversarial setting against a particular model \cite{bartolo-etal-2020-beat}. However, these methods can be expensive and difficult to scale.

This has led to an increased interest in creating \emph{automatic} counterfactual data for evaluating out-of-distribution generalization \cite{bowman-dahl-2021-will} and for counterfactual data augmentation \cite{geva2021break,longpre2021entity}. Some work focuses on using heuristics like 
swapping superlatives and nouns \cite{dua2021learning}, changing gendered words \cite{webster2020measuring}, or targeting specific data splits \cite{finegan-dollak-verma-2020-layout}. 
More recent work has focused on using meaning representation frameworks and structured control codes \citep{wu-etal-2021-polyjuice}, including grammar formalisms \cite{li2020linguistically}, semantic role labeling \cite{ross2021tailor}, structured image representations like scene graphs \cite{bitton2021automatic}, and query decompositions in multi-hop reasoning datasets \cite{geva2021break}. \citet{ye2021evaluating} and \citet{longpre2021entity} perturb contexts instead of questions by swapping out all mentions of a named entity. 
The change in label can be derived heuristically 
or requires a round of human re-labeling of the data.
These may also be difficult to apply to tasks like Natural Questions \citep{kwiatkowski2019natural}, where pre-defined schemas can have difficulty covering the range of 
semantic perturbations that may be of interest.

\subsection{Data Augmentation}

Non-counterfactual data augmentation methods for QA, where the synthetic examples are \emph{not} paired with the original data, have shown only weak improvements to robustness and out-of-domain generalization \cite{bartolo2021improving, lewis2021paq}. Counterfactual data augmentation is hypothesized to perform better, as exposing the model to minimal pairs should reduce spurious correlations and make the model more likely to learn the correct, causal features \citep{kaushik2019learning}. However,
\citet{joshi2021investigation} find that methods that limit the structural and semantic space of perturbations can potentially hurt generalization to other types of transformations.
This problem is exacerbated in the question answering scenario where there can be multiple semantic dimensions to edit. 
Our method attempts to address this by targeting a broad range of semantic phenomena, thus reducing the chance for the augmented model to overfit.

\section{RGF: Counterfactuals for Information-seeking Queries}
\label{sec:experimental}

We define a counterfactual example as an alternative input $x'$ which differs in some meaningful, controlled way from the original $x$, which in turn allows us to reason -- or teach the model -- about changes in the label (the outcome). 
For question-answering, we take as input triples $(q, c, a)$ consisting of the question, context passage, and short answer, and produce counterfactual triples $(q', c', a')$ where $a' \ne a$.
This setting poses some unique challenges, such as the need for background knowledge to identify relevant semantic variables to alter, ensuring sufficient semantic diversity in question edits, and avoiding questions with false premises or no viable answers. Ensuring (or characterizing) minimality can also be a challenge, as small changes to surface form can lead to significant semantic changes, and vice-versa. We introduce a general paradigm for data generation --- \emph{\textbf{R}etrieve, \textbf{G}enerate and \textbf{F}ilter} --- to tackle these challenges.

\subsection{Overview of RGF}
\label{sec:rgf-overview}

An outline of the RGF method is given in Figure~\ref{fig:teaser}. Given an input example $x = (q, c, a)$ consisting of a question, a context paragraph, and the corresponding answer, RGF generates a set of new examples $N(x) = \{(q_1', c_1', a_1'), (q_2', c_2', a_2'), \dots \}$
from the local neighborhood around x.
We first use an open-domain retrieve-and-read model
to retrieve alternate contexts $c'$ and answers $a'$ where $a \neq a'$. As near-misses for a task model, these candidates $(c', a')$ are closely related to the original target $(c, a)$ but often differ along interesting, latent semantic dimensions (Figure~\ref{fig:neighborhood}) in their relation to the original question, context, and answer.
We then use a sequence-to-sequence model 
to generate new questions $q'$ from the context and answer candidates $(c', a')$. This yields triples $(q',c',a')$ which are fully labeled, 
avoiding the problem of unanswerable or false-premise questions. 

Compared to methods that rely on a curated set of minimal edits \citep[e.g.][]{wu2021polyjuice,ross2021tailor}, our method admits the use of alternative contexts\footnote{An alternative approach would be to make direct, targeted edits to the original context $c$. However, beyond a limited space of local substitutions \citep{longpre2021entity,ye2021evaluating, ross-etal-2021-explaining} this is very difficult due to the need to model complex discourse and knowledge relations.} $c' \ne c$, and we do not explicitly constrain our triples to be \textit{minimal} perturbations during the generation step.
Instead, we use post-hoc filtering to reduce noise, select minimal candidates, or select for specific semantic phenomena based on the relation between $q$ and $q'$. This allows us to explore a significantly more diverse set of counterfactual questions $q'$ (\S\ref{sec:additional_intrinsic}), capturing relations that may not be represented in the original context $c$.

We describe each component of RGF below; additional implementation details are provided in Appendix~\ref{sec:implimentation_details}.

\subsection{Retrieval}
\label{sec:rgf-retrieval}

We use REALM retrieve-and-read model of \cite{guu2020realm}. REALM consists of a BERT-based bi-encoder for dense retrieval, a dense index of Wikipedia passages, and a BERT-based answer-span extraction model for reading comprehension, all fine-tuned on Natural Questions \citep[NQ;][]{kwiatkowski2019natural}. Given a question $q$, REALM outputs a ranked list of contexts and answers within those contexts: $\{ (c'_1,a'_1), (c'_2, a'_2), \dots (c'_k, a'_k) \}$. 
These alternate contexts and answers provide relevant yet diverse background information to construct counterfactual questions. For instance, in Figure \ref{fig:teaser}, the question \textit{``Who is the captain of the Richmond Football Club"} with answer \textit{``Trent Cotchin"} also returns other contexts with alternate answers like \textit{``Jeff Hogg"} ($q'=$\textit{``Who captained the team in 1994"}),
and \textit{``Steve Morris"} ($q'=$\textit{``Who captained the reserve team in the VFL league"}). 
Retrieved contexts can also capture information about closely related or ambiguous entities.
For instance, the question \textit{``who wrote the treasure of the sierra madre"} retrieves passages about the original book \textit{Sierra Madre}, its movie adaptation, and a battle 
fought in the Sierra de las Cruces mountains. This background knowledge allows us to perform \emph{contextualized} counterfactual generation, without needing to specify \textit{a priori} the type of perturbation or semantic dimension. To focus on label-transforming counterfactuals, we retain all $( c'_i, a'_i )$ where $a'_i$ does not match any of the gold answers $a$ from the original NQ example.

\subsection{Question Generation}
\label{sec:rgf-qgen}
This component generates questions $q'$ that correspond to the answer-context pairs $(c', a')$. We use a T5 \citep{raffel2019exploring} model fine-tuned on $(q,c,a)$ triples from Natural Questions, using context passages as input with the answer marked with special tokens.
We use the trained model to generate questions $( q'_1, q'_2, \dots q'_k )$ for each of the the retrieved set of alternate contexts and answers, $( (c'_1,a'_1), (c'_2, a'_2), \dots (c'_k, a'_k))$. For each $(c'_i,a'_i)$, we use beam decoding to generate 15 different questions $q'$.
We measure the fluency and correctness of generated questions in \S\ref{sec:intrinsic}.

\subsection{Filtering for Data Augmentation}
\label{sec:rgf-filtering}

\paragraph{Noise Filtering}
The question generation model can be noisy, resulting in a question that cannot be answered given $c'$ or for which $a'$ is an incorrect answer. Round-trip consistency \cite{alberti2019synthetic, fang2020accelerating} uses an existing QA model to answer the generated questions, ensuring that the
predicted answer is consistent with the target answer provided to the question generator. 
We use an ensemble of six T5-based reading-comprehension ($(q,c) \to a$) models, trained on NQ using different random seeds (Appendix~\ref{sec:implimentation_details}), and keep any generated $(q', c', a')$ triples where at least 5 of the 6 models agree on the answer.
This discards about 5\% of the generated data, although some noise still remains; see \S\ref{sec:intrinsic} for further discussion.

\paragraph{Filtering for Minimality}
Unlike prior work on generating counterfactual perturbations, we do not explicitly control for the type of semantic shift or perturbation in the generated questions. 
Instead, we use post-hoc filtering over generated questions $q'$ to encourage minimality of perturbation. We define a filtering function $f(q, q')$ that categorizes the semantic shift or perturbation in $q'$ with respect to $q$. One simple version of $f$ is the word-level edit (Levenshtein) distance between $q$ and $q'$. 
After 
noise filtering, for each original $(q,c,a)$ triple we select the generated $(q', c', a')$ with the smallest non-zero word-edit distance between $q$ and $q'$ such that $a \neq a'$. 
We use this simple heuristic to create large-scale \emph{counterfactual training data} for augmentation experiments (\S\ref{sec:data_augmenation}). Over-generating potential counterfactuals based on latent dimensions identified in retrieval and using a simple filtering heuristic avoids biasing the model toward a narrow set of perturbation types \cite{joshi2021investigation}.

\begin{figure*}[ht!]
    \centering
    \includegraphics[scale=0.58]{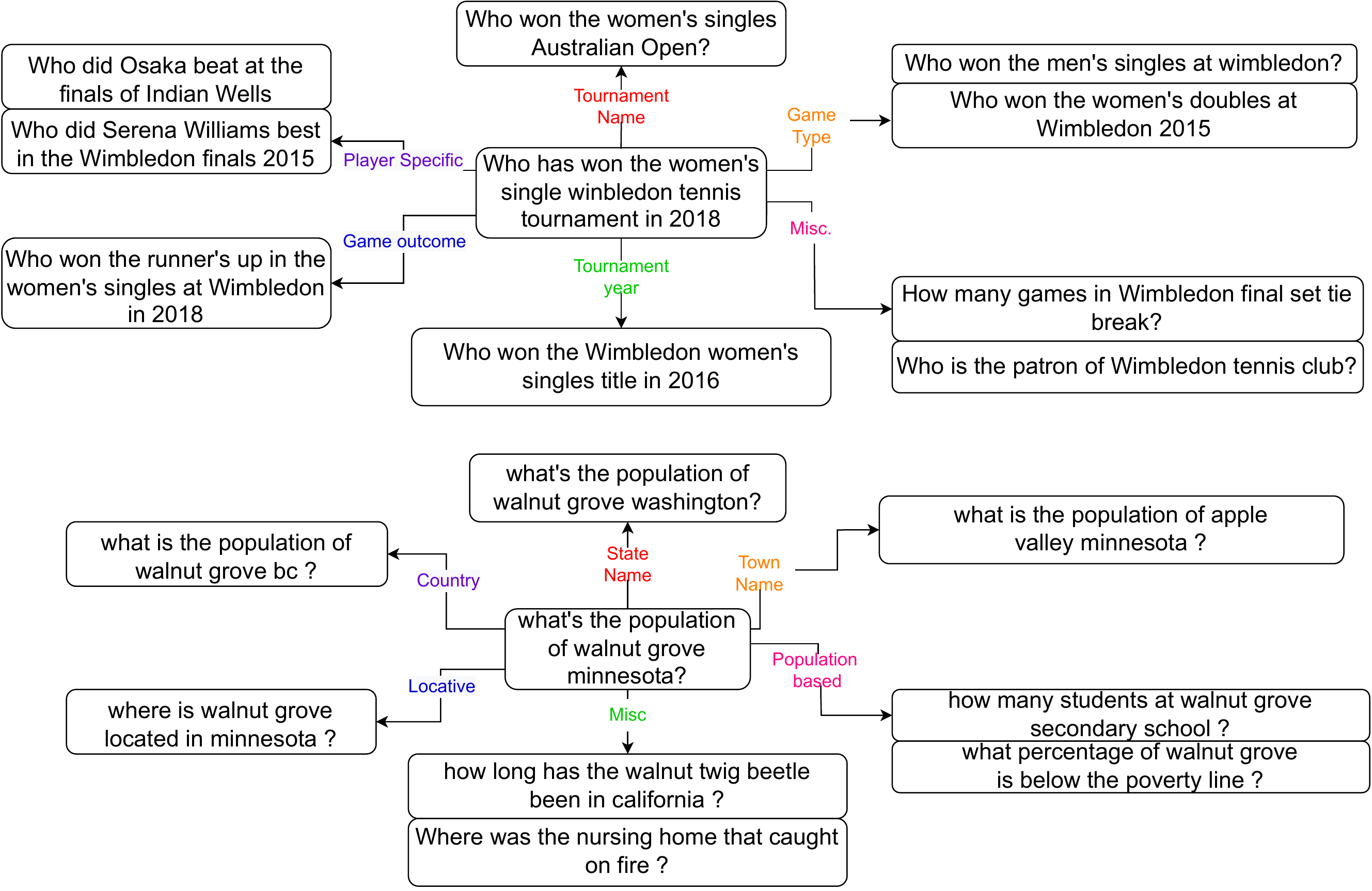}
    \caption{Context-specific semantic diversity of perturbations achieved by RGF on questions from NQ. The multiple latent semantic dimensions identified (arrows in the diagram) emerge from our retrieval-guided approach.}
    \label{fig:neighborhood}
\end{figure*}

\subsection{Semantic Filtering for Evaluation}
\label{sec:filtering-eval}

To better understand the types of counterfactuals generated by RGF, we can apply additional filters based on question meaning representations to categorize counterfactual $(q, q')$ pairs for evaluation.
Meaning representations provide a way to decompose a question into semantic units and categorize $(q, q')$ based on which of these units are perturbed.   
In this work, we employ the QED formalism for explanations in question answering \cite{lamm2020qed}.
QED decompositions segment the question into a predicate template and a set of reference phrases. For example, the question \textit{``Who is captain of richmond football club"} decomposes into one question reference \textit{``richmond football club"} and the predicate \textit{``Who is captain of X"}. A few example questions and their  QED decompositions are illustrated in Table~\ref{tab:qed_summary}.

\begin{table}[t!]
    \centering
    \small
    \begin{tabular}{@{}l@{}}
    \toprule
    \makecell[l]{\textbf{Question from NQ} \\ 
        \textbf{Original:} who is the captain of \hlc[CornflowerBlue]{richmond football club}?\\
        \textbf{Predicate:} who is the captain of X?} \\ \midrule
    \makecell[l]{\textbf{Reference Change} \\
        \textbf{CF1:} who is the captain of \hlc[LimeGreen]{richmond's vfl reserve team}? \\
        \textbf{Predicate:} who is the captain of X?} \\[0.5cm] 
    \makecell[l]{\textbf{Predicate Change} \\ 
        \textbf{CF2:} who wears \hlc[Gray]{number 9} for \hlc[CornflowerBlue]{richmond football club}? \\
        \textbf{Predicate:} who wears Y for X?} \\[0.5cm]
    \makecell[l]{\textbf{Predicate and Reference Change} \\ 
        \textbf{CF3:} who did \hlc[Yellow]{graham} negate in \hlc[VioletRed]{the grand final} last year? \\
        \textbf{Predicate:} who did X negate in Y last year?} \\ \bottomrule
    \end{tabular}
    \caption{Categorization of generated questions based on QED decomposition.
    The original reference \textit{``Richmond football Club"} changes in CF1 and CF3. Predicate \textit{``Who is the captain"} changes in CF2 and CF3.}
    \label{tab:qed_summary}
\end{table}

We use these question decompositions to identify the relation between a counterfactual pair $(q,q')$.
Concretely, we fine-tune a T5-based model on the QED dataset to perform explanation generation following the recipe of \citet{lamm2020qed}, and use this to identify predicates and references for the question from each $(q,c,a)$ triple.
We use exact match between strings to identify reference changes. 
As predicates can often differ slightly in phrasing (\textit{who captained} vs. \textit{who is captain}), we take a predicate match to be a prefix matching with more than 10 characters.
For instance, \textit{``Who is the captain of Richmond's first ever women's team?"}, \textit{``Who is the captain of the Richmond Football Club"} have same predicates.
We filter generated questions into three perturbation  categories --- reference change, predicate change, or both.

\section{Intrinsic Evaluation}
\label{sec:intrinsic}

Following desiderata from \citet{wu-etal-2021-polyjuice} and \citet{ross2021tailor}, we evaluate our RGF data along three 
measures: \emph{fluency}, \emph{correctness}, 
 and \emph{directionality}.

\begin{table}[t!]
    \centering
    \small
    \begin{tabular}{l|l}
    \toprule
      Semantic Change &  Example (\textcolor{blue}{Original}, \textcolor{red}{Counterfactual}) \\
    \midrule
    Reference Change & \textcolor{blue}{O}: when did lebron\_james join \\
    TAILOR  &  the \underline{Miami\_Heat}?  \textcolor{red}{C}: When did  \\
    \cite{ross2021tailor} & lebron\_james come into the \underline{league}? \\
    \midrule 
    Predicate Change  & \textcolor{blue}{O}: Who \underline{won} the war between india \\
    AmbigQA  & and pakistan %
    \textcolor{red}{C}: Who \underline{started}  \\
    \cite{min-etal-2020-ambigqa} & the war between india and pakistan \\ %
    \midrule
    Disambiguation  & \textcolor{blue}{O}: When does walking dead season\\ 
    AmbigQA &  8 start?  \textcolor{red}{C}: When does walking  \\
    \cite{min-etal-2020-ambigqa} & dead season 8 \underline{second half} start?\\
    \midrule
    Negation &  \textcolor{blue}{O}: what religion \underline{observes} the \\
    Polyjuice & sabbath day  \textcolor{red}{C}: what religion \underline{does} \\
    \cite{wu-etal-2021-polyjuice} & \underline{not keep} the sabbath day \\
    \bottomrule
    \end{tabular}
    \caption{Patterns of semantic change between original queries (O) and RGF counterfactuals (C), corresponding to patterns explored by related works.}
    \label{tab:qualitative}
\end{table}

\paragraph{Fluency}
Fluency measures whether the generated text is grammatically correct and semantically meaningful. Fluency is very high from RGF, as the generation step leverages a high-quality pretrained langauge model (T5).
We manually annotate a subset of 100 generated questions, and find that 96\% of these are fluent.

\paragraph{Correctness}
Correctness measures if the generated question $q'$ and context, alternate answer pairs $(c',a')$ are aligned, i.e. the question is answerable given context $c'$ and $a'$ is that answer.  
We quantify correctness in the generated dataset by manually annotating a samples of 100 $( q', c', a' )$ triples (see Appendix~\ref{sec:appendix_human}). The proportion of noise varies from 30\% before noise filtering and 25\% after noise filtering using an ensemble of models (\S\ref{sec:rgf-filtering}).

\paragraph{Directionality/Semantic Diversity}

In Table \ref{tab:qualitative}, we show examples of semantic changes that occur in our data, including reference changes (50\% of changes), predicate changes (30\%), negations (1\%), question expansions, disambiguations, and  contractions (13\%). These cover many of the transformations found in prior work \cite{gardner-etal-2020-evaluating, ross2021tailor, min-etal-2020-ambigqa}, but RGF is able to achieve these without the use of heuristic transformations or structured meaning representations.
As shown in Figure~\ref{fig:neighborhood}, the types of relations are semantically rich and cover attributes relevant to each particular instance that would be difficult to capture with a globally-specified schema. Additional examples are shown in  Figure~\ref{fig:neighborhood_additional_examples}.

\begin{table*}[ht!]
    \centering
    \small
    \begin{tabular}{c|c|c|c|c|c|c|c|c}
    \toprule
     \bf Exact Match (RC) & \bf Train Size & \multicolumn{1}{c|}{\bf NQ}  & \multicolumn{1}{c|}{\bf SQuAD} & \multicolumn{1}{c|}{\bf\small TriviaQA} & \multicolumn{1}{c|}{\bf\small HotpotQA} & \multicolumn{1}{c|}{\bf\small BioASQ} & \multicolumn{1}{c|}{\bf\small AQA} & \multicolumn{1}{c}{\bf\small AmbigQA} \\ 
     \midrule
     Original NQ
     & 90K 
& 70.91 %
& 80.26 %
& 13.67 %
& 50.57 %
& 35.90 %

& 27.00 %
& 46.81 \\ %
Ensemble
& 90K 
& \bf 71.29 %
& \bf 80.50 %
& 13.86 %
& 50.57 %
& 36.90 %
& 27.80
& \bf 46.90 \\ %
Gold Agen-Qgen 
& 90K + 90K
& 70.80
& 67.71 %
& 10.83 %
& 42.69 %
& 30.63 %

& 19.40 %
& 41.95 \\
Rand. Agen-Qgen 
& 90K + 90K
& 71.06 %
& 74.31 %
& 12.88 %
& 45.52 %
& 32.58 %

& 23.30 %
& 42.48 \\
    \midrule
RGF (REALM-Qgen) 
& 90K + 90K
&     70.22 %
& 79.87

& \bf 15.39 %
& \bf 53.36
& \bf 42.89 %

& \bf 28.90
& 46.81 \\ %
    \bottomrule
    \end{tabular}
    \caption{Exact Match results for the reading comprehension task for in-domain NQ development set, out-of-domain datasets from MRQA 2019 Challenge \cite{fisch2019mrqa}, Adversarial QA \cite{bartolo-etal-2020-beat} and AmbigQA \cite{min-etal-2020-ambigqa}. RGF improves out-of-domain and challenge-set performance compared to other data augmentation baselines.}
    \label{tab:rc_ood}
\end{table*}

\section{Data Augmentation}
\label{sec:data_augmenation}
\vspace{-5pt}
Unlike many counterfactual generation methods, RGF natively creates fully-labeled $(q', c', a')$ examples which can be used directly for counterfactual data augmentation (CDA). We augment the original NQ training set with additional examples from RGF, shuffling all examples in training. We explore two experimental settings, reading comprehension (\S\ref{results:rc}) and open-domain QA (\S\ref{results:odqa}), and compare RGF-augmented models to those trained only on NQ, as well as to alternative baselines for synthetic data generation. As described in Section~\ref{sec:rgf-filtering}, we use edit-distance based filtering to choose \emph{one} generated $(q', c', a')$ triple to augment for every original example, $(q,c,a)$.\footnote{We don't see significant gains from adding more data beyond this; see Appendix~\ref{results:rc-additional-training-data}}
Additional training details for all models and baselines are included in Appendix~\ref{sec:implimentation_details}.

\subsection{Baselines}
\label{sec:baselines}

In the abstract, our model for generating counterfactuals specifies a way of selecting contexts $c'$ from original questions, and answers $a'$ within those contexts, and a way of a generating questions $q'$ from them. RGF uses a retrieval model to identify relevant contexts; here we experiment with two baselines that use alternate ways to select $c'$. We also compare to the \textbf{ensemble} of six reading comprehension models described in \ref{sec:rgf-filtering}, with answers selected by majority vote.

\paragraph{Random Passage (Rand. Agen-Qgen)}
Here, $c'$ is a randomly chosen paragraph from the Wikipedia index, with no explicit relation with the original question. This setting simulates generation from the original data distribution of Natural Questions. 
To ensure that the random sampling of Wikipedia paragraphs has a similar distribution, we employ the learned passage selection model from \citet{lewis2021paq},\footnote{\url{https://github.com/facebookresearch/PAQ}}.
This baseline corresponds to the model of \citet{bartolo2021improving}, which was applied to the SQuAD dataset \citep{rajpurkar2016squad}; our version is trained on NQ and omits AdversarialQA.

\paragraph{Gold Context (Gold Agen-Qgen)}
Here, $c'$ is 
the passage $c$  containing the original short answer $a$ 
from the NQ training set. 
This baseline specifically ablates the retrieval component of RGF, testing whether the use of alternate passages leads to more diversity in the resulting counterfactual questions.

\paragraph{Answer Generation for Baselines}
For both the above baselines for context selection, we 
select spans in the new passage that are likely to be answers for a potential counterfactual question. 
We use a T5 \citep{raffel2019exploring} model fine-tuned for question-independent answer selection $c \to a$ on NQ, and select the top 15 candidates from beam search.
To avoid simply repeating the original question, we only retain answer candidates $a'$ which do not match the original NQ answers $a$ for that example. 
These alternate generated answer candidates and associated passages are then used for question generation and filtering as in RGF (\S\ref{sec:rgf-qgen}). For the Gold Agen-Qgen case, we select based on the longest edit distance between $(q,q')$, which gave significantly better performance than random selection or the shortest edit distance used for RGF.

\subsection{Reading Comprehension (RC)}\label{results:rc}

In the reading comprehension (RC) setting, the input consists of the question and context and the task is to identify an answer span in the context. Thus, we augment training with full triples $( q',c',a' )$ consisting of the retrieved passage $c'$, generated and filtered question $q'$, and alternate answer $a'$.

\paragraph{Experimental Setting}
We finetune a T5 \citep{raffel2019exploring} model for reading comprehension, with input consisting of the question prepended to the context.
We evaluate domain generalisation of our RC models on three evaluation sets from the MRQA 2019 Challenge \cite{fisch2019mrqa}. We also measure performance on evaluation sets consisting of counterfactual or perturbed versions of RC datasets on Wikipedia, including SQuAD \citep{rajpurkar2016squad}, AQA \citep[adversarially-generated SQuAD questions;][]{bartolo-etal-2020-beat},
and human authored counterfactual examples \citep[contrast sets;][]{gardner-etal-2020-evaluating} from the QUOREF dataset \cite{dasigi2019quoref}. We also evaluate on the set of disambiguated queries in AmbigQA \cite{min-etal-2020-ambigqa}, which by construction are minimal edits to queries from the original NQ.

\begin{table*}[ht!]
    \centering
    \begin{tabular}{c|c|c|c|c|c|c}
        \toprule
        \bf Exact Match (OD) & \bf Train Size & \bf \small NQ & \bf \small TriviaQA & \bf \small AmbigQA & \bf \small SQuAD v1.0 & \bf \small TREC \\
        \midrule
        Original &	90K	&	37.65	& 26.75 & 	22.43	& 14.25	& 31.93 \\ 
        Gold Agen-Qgen  &	90K + 90K	& 37.86	& 27.02	& 23.65	& 15.01	& 32.94 \\ 
        Rand. Agen-Qgen	& 90K + 90K 	& 37.45 &	29.87 &	 24.13 &	14.55 &	 26.89 \\
        \midrule
        RGF (REALM-Qgen) & 	90K + 90K	& \bf 39.11 &	\bf 32.32 &	\bf 26.98	& \bf 16.94 &		\bf 33.61 \\
        \bottomrule
    \end{tabular}
    \caption{Exact Match results on open-domain QA datasets (TriviaQA, AmbigQA, SQuAD and TREC) for data augmentation with RGF counterfactuals and baselines. Open-domain improvements are larger than in the RC setting, perhaps as the more difficult task benefits more from additional data.}
    \label{tab:opendomain}
\end{table*}

\vspace{-5pt}
\paragraph{Results}
We report exact-match scores in Table~\ref{tab:rc_ood}; F1 scores follow a similar trend. We observe only limited improvements on the in-domain NQ development set, but we see significant improvements from CDA with RGF data in out-of-domain and challenge-set evaluations compared both to the original NQ model and the Gold and Random baselines.
RGF improves by 1-2 EM points on most challenge sets, and up to 7 EM points on the BioASQ set compared to training on NQ only, while baselines often underperform the NQ-only model on these sets.
Note that all three augmentation methods have similar proportion of noise (Appendix \ref{sec:appendix_human}), so CDA's benefits may be attributed to improving model's ability to learn more robust features for the task of reading comprehension. Using an ensemble of RC models improves slightly on some tasks, but does not improve on OOD performance as much as RGF.
RGF's superior performance compared to the Gold Agen-Qgen baseline is especially interesting, since the latter also generates topically related questions. We observe that RGF counterfactuals are more closely related to the original question compared to this baseline (Figure \ref{fig:question_distribution} in Appendix~\ref{sec:additional_results}), since $q'$ is derived from a near-miss candidate $(c', a')$ to answer the original $q$ (S\ref{sec:rgf-overview}).

\subsection{Open-Domain Question Answering (OD)}\label{results:odqa}

In the open-domain (OD) setting, only the question is provided as input. The pair $( q',a' )$, consisting of generated and filtered question $q'$ and alternate answer $a'$, is used for augmentation. Compared to the RC setting where passages change as well, here the edit distance filtering of \S\ref{sec:rgf-filtering} ensures the augmentation data represents minimal perturbations.

\paragraph{Experimental Setting}
We use the method and implementation from \citet{guu2020realm} to finetune REALM on $(q,a)$ pairs from NQ. 
End-to-end training of REALM updates both the reader model and the query-document encoders of the retriever module. 
We evaluate domain generalization on popular open-domain benchmarks: TriviaQA \cite{joshi2017triviaqa}, SQuAD \cite{rajpurkar2016squad}, Curated TREC dataset \cite{min2021joint}, and disambiguated queries from AmbigQA \cite{min-etal-2020-ambigqa}.

\begin{table*}[ht!]
    \centering
    \begin{tabular}{c|c|c|c|c|c|c}
        \toprule
        \bf Consistency (RC) & \bf Train Size & \bf \small AQA & \bf \small AmbigQA & \bf \small QUOREF-C & \bf \small RGF ($\Delta$ Ref)  & \bf \small RGF  ($\Delta$ Pred) \\
        \midrule
        Original NQ
        & 90K 
        & 63.22 &	51.72&	44.86&	64.65&	52.93 \\
        Ensemble 
        & 90K
        & 63.87 &	48.33 &	46.02 &	65.21 &	55.21 \\
        Gold Agen-Qgen
        & 90K + 90K 
        & 50.25	& 42.86	& 40.66	& 55.63 &	43.08 \\
        Rand. Agen-Qgen
        & 90K + 90K 
        & 56.07 &	48.08 &	44.79 &	60.06 &	48.34 \\
        \midrule
        RGF (REALM-Qgen) 
        & 90K + 90K 
        & \bf 64.46 &	\bf 55.93 &	\bf 48.94 &	76.17 &	\bf 66.12 \\
        \midrule
        \midrule
        RGF $\Delta$ Ref.
        & 90K + 52K 
        & 58.8 &	56.9 &	40.54 &	\bf 77.61 &	59.56 \\
                RGF $\Delta$ Pred. 
        & 90K + 52K 
        & 63.64 &	49.15 &	43.13 &	73.29 &	63.09 \\
        \bottomrule
    \end{tabular}
    \caption{Results for pairwise consistency (\S\ref{results:robust}) on reading comprehension, measured for datasets containing pairs of very similar questions. QUOREF-C refers to the QUOREF contrast set from \citep{gardner-etal-2020-evaluating}. RGF leads to better consistency in RC and open-domain settings (Appendix \ref{results:odqa-consistency}). Results on effect of perturbation type during training ($\Delta$ Ref. and $\Delta$ Pred.) suggest that perturbation-bias does not degrade consistency over the original model.}
    \label{tab:consistency}
\end{table*}

\paragraph{Results}
In the open-domain setting (Table \ref{tab:opendomain}), we observe an improvement of 2 EM points over the original model even in-domain on Natural Questions, while also improving significantly when compared to other data augmentation techniques. RGF improves over the next best baseline --- Random Agen-Qgen --- by up to 6 EM points (on TriviaQA).
We hypothesize that data augmentation has more benefit in this setting, as the open-domain task is more difficult than reading comprehension, and counterfactual queries may help the model learn better query and document representations to improve retrieval.

\section{Analysis}
\label{sec:analysis}
To better understand how CDA improves the model, we introduce a measure of local consistency (\S\ref{results:robust}) to measure model robustness, and perform a stratified analysis (\S\ref{results:bias}) to show the benefits of the semantic diversity available from RGF. 

\subsection{Local Robustness}
\label{results:robust}

Compared to synthetic data methods such as PAQ \citep{lewis2021paq}, RGF generates counterfactual examples that are paired with the original inputs and concentrated in local neighborhoods around them (Figure~\ref{fig:neighborhood}). As such, we hypothesize that augmentation with this data should specifically improve local consistency, i.e. how the model behaves under small perturbations of the input.

\paragraph{Experimental Setting}
We explicitly measure how well a model's local behavior respects perturbations to input. Specifically, if a model $f : (q,c) \to a$ correctly answers $q$, how often does it also correctly answer $q'$? We define \textit{pairwise consistency} as accuracy over the counterfactuals $(q', a', c')$, conditioned on correct predictions for the original examples:
$$ \mathbb{C}(D) = E_D[f(q', c') = a'\ |\ f(q, c) = a] $$
To measure consistency, we construct validation sets consisting of paired examples $(q,c,a), (q',c',a')$: one original, and one counterfactual. We use QED to categorize our data, as described in \S\ref{sec:filtering-eval}. Specifically, we create two types of pairs: (a) a change in reference where question predicate remains fixed, and (b) a change in predicate, where the original reference(s) are preserved.\footnote{We require that the new reference set is a superset of the original, since predicate changes can introduce additional reference slots (see CF2 in Table~\ref{tab:qed_summary}).}
We create a clean evaluation set by first selecting RGF examples for predicate or reference change, then manually filtering the data to discard incorrect triples (\S\ref{sec:intrinsic}) until we have 1000 evaluation pairs of each type (see Appendix~\ref{sec:appendix_human}).

We also construct paired versions of AQA, AmbigQA, and the QUOREF contrast set. For AmbigQA, we pair two disambiguated questions and for QUOREF, we pair original and human-authored counterfactuals. 
AQA consists of human-authored adversarial questions $q'$ which are not explicitly paired with original questions; we create pairs by randomly selecting an original question $q$ and a generated question $q'$ from the same passage.
\paragraph{Results}
Training with RGF data improves consistency by 12-14 points on the QED-filtered slices of RGF data,
and 5-7 points on AQA, AmbigQA and QUOREF contrast (Table \ref{tab:consistency}).
The Gold Agen-Qgen baseline (which contains topically related queries about the same passage) also improves consistency over the original model compared to the Random Agen-Qgen baseline or to the ensemble model, though not by as much as RGF. 
Consistency improvements on AQA, AmbigQA and QUOREF are especially noteworthy, since they suggest an improvement in robustness to local perturbations that is independent of other confounding distributional similarities between training and evaluation data.

\subsection{Effect of Perturbation Type}
\label{results:bias}

QED-based decomposition of queries allows for the creation of label-changing counterfactuals along  \emph{orthogonal} dimensions --- a change of reference or predicate. %
We investigate whether training towards one type of change induces generalization bias, a detrimental effect which has been observed in tasks such as NLI \cite{joshi2021investigation}.

\paragraph{Experimental Setting}
We shard training examples into two categories based on whether $q$ and $q'$ have the same reference (predicate change) or same predicate (reference change), as defined in \S\ref{sec:filtering-eval}. We over-generate by starting with 20 $(q', c', a')$ for each original training example to ensure that we find at least one $q'$ that matches the criterion.
We also  evaluate on paired evaluation sets from \S\ref{results:robust}.

\paragraph{Results}
Results are shown for QED-filtered training in Table~\ref{tab:consistency}. 
Counterfactual perturbation of a specific kind (a predicate or a reference change) during augmentation does not hurt performance on another perturbation type compared to the baseline NQ model, which differs from the observations of \citet{joshi2021investigation} on NLI. Furthermore, 
similar to the observations of \citet{min2020syntactic},
augmenting with one type of perturbation has
orthogonal benefits that improve model generalization on another perturbation type: augmenting with RGF ($\Delta$ Pred.) leads to significant improvement on RGF ($\Delta$ Ref.), and vice-versa.
Compared to reference-change examples, augmenting with predicate-change examples leads to greater improvements in local consistency,
except for on RGF ($\Delta$ Ref.) and on AmbigQA -- which contains many reference-change pairs.
Predicate-change examples may also be more informative to the model, as reference changes can be modeled more easily by lexical matching within common context patterns.

\subsection{Effect of Training data size}\label{results:low}

\citet{joshi2021investigation} show CDA to be most effective in the low-resource regime. To better understand the role that dataset size plays in CDA in the reading comprehension setting, we evaluate RGF in a cross-domain setting where only a small amount of training data is available.

\paragraph{Experimental Setting} 
Since our approach depends on using an open-domain QA model and a question generation model trained on all Natural Questions data, we instead experiment with a low-resource transfer setting on the BioASQ domain, which consists of questions on the biomedical domain. We use the domain-targeted retrieval model from \citet{ma2021zero}, where synthetic question-passage relevance
pairs generated over the PubMed corpus are used to train domain-specific retrieval without any gold supervised data. We fine-tune our question-generation model on (limited) in-domain data, generate RGF data for augmentation, and then use this along with (limited) in-domain data to further fine-tune an RC model, using the NQ-trained weights for initialization. Further training details are provided in Appendix~\ref{sec:implimentation_details}.

\begin{table}[ht]
    \centering
    \begin{tabular}{l|l|cc}
        \toprule
        \bf Training Data & \bf Train Size & \multicolumn{2}{c}{\bf BioASQ (Dev)} \\
        & & F1 & EM \\
        \midrule
        Original & 1000 & 42.93 & 23.67 \\
        Orig. + RGF & 500 + 500 & 41.72 & 23.01 \\
        \midrule
        Original & 2000 & \bf 45.88 & 25.80 \\
        Orig. + RGF & 1000 + 1000 & 44.64 & \bf 26.80 \\
        \bottomrule
    \end{tabular}
    \caption{Results on the reading comprehension task for Low Resource Transfer setting on BioASQ 2019 dataset. A model trained on 1000 gold BioASQ plus 1000 RGF examples performs nearly as well as a model trained on 2000 gold examples.}
    \label{tab:low}
\end{table}

\paragraph{Results}
We observe significant improvements over the baseline model in the low resource setting for in-domain data (< 2000 examples), as shown in Table~\ref{tab:low}. Compared with the limited gains we see on the relatively high-resource NQ reading comprehension task, we find that on BioASQ, CDA with 1000 examples improves performance by 2\% F1 and 3\% exact match, performing nearly as well as a model trained on 2000 gold examples. 
These results suggest that using counterfactual data in lieu of collecting additional training data is especially useful in the low-resource setting.

\section{Conclusion}
Retrieve-Generate-Filter (RGF) creates counterfactual examples for QA which are semantically diverse, using knowledge from the passage context and a retrieval model to capture semantic changes that would be difficult to specify \textit{a priori} with a global schema. The resulting examples are fully-labeled, and can be used directly for training or filtered using meaning representations for analysis.
We show that training with this data leads to improvements on open-domain QA, as well as on challenge sets, and leads to significant improvements in local robustness.
While we focus on question answering, for which retrieval components are readily available, we note that the RGF paradigm is quite general and could potentially be applied to other tasks with a suitable retrieval system.

\bibliography{anthology,custom}
\bibliographystyle{acl_natbib}

\clearpage
\appendix

\section{Model Training and Implementation Details}
\label{sec:implimentation_details}

Below, we describe the details of different models trained in the RGF pipeline. Unless specified otherwise, we use the T5X library\footnote{\url{https://github.com/google-research/t5x}} and pre-trained checkpoints from \citet{raffel2019exploring}\footnote{ \url{https://github.com/google-research/text-to-text-transfer-transformer\#released-model-checkpoints}}.

\paragraph{Question Generation}
We use a T5-3B model fine-tuned on Natural Questions (NQ) dataset. We only train on the portion of the dataset that consists of gold short answers and an accompanying long answer evidence paragraph from Wikipedia.
The input consists of the title of the Wikipedia article the passage is taken from, a separator (`>>') and the passage. The short answer is enclosed in the passage using character sequences `« answer =' and `»' on left and right respectively. The output is the original NQ question. The input and output sequence lengths are restricted to be $640$ and $256$ respectively.  
We train the model for $20k$ steps with a learning rate of $2\cdot 10^{-5}$, dropout $0.1$, and batch size of $128$. We decode with a beam size of $15$, and take the top candidate as our generated question $q'$.

\paragraph{Answer Generation}
We use a T5-3B model trained on the same subset of Natural Questions (NQ) as question generation with same set of hyperparameters and model size described above. The input consists of the title of the Wikipedia article the passage is taken from, a separator (`>>') and the passage, while the output sequence is the short answer from NQ.

\paragraph{Reading Comprehension Model}
We model the task of span selection-based reading comprehension, i.e. identifying an answer span given question and passage, as a sequence-to-sequence problem. Input consists of the question, separator (`>>'), and title of Wikipedia article, separator (`>>') and passage. The answer format is simply one of the gold answer strings.
The reading comprehension model is a T5-large model trained with batch size of $512$ and learning rate $2\cdot 10^{-4}$  for $20K$ steps. 

\paragraph{Open-domain Question Answering model}
The open domain QA model is based on the implementation from \citet{lee2019latent}, and initialized with the REALM checkpoint from \citet{guu2020realm}\footnote{\url{https://github.com/google-research/language/tree/master/language/realm}}. Both the retriever and reader are initialized from the BERT-base-uncased model. The query and document representations are 128 dimensional vectors. When finetuning, we use a learning rate of $10^{-5}$
and a batch size of 1 on a single Nvidia V100 GPU. We perform 2 epochs of fine-tuning for Natural Questions.

\paragraph{Noise Filtering}
We train 6 reading comprehension models based on the configurations above with different seed values for randomizing training dataset shuffling and optimizer initialization. We retain examples where more than 5 out of 6 models have the same answer for a question.

\paragraph{QED Training}
We use a T5-large model fine-tuned on the Natural Questions subset with QED annotations \cite{lamm2020qed}.\footnote{\url{https://github.com/google-research-datasets/QED}} We refer the reader to the QED paper for details on the linearization of explanations and inputs in the T5 model.
Our model is fine-tuned with batch size of $512$ and learning rate $2\cdot 10^{-4}$ for 20k steps.

\paragraph{Experimental Variability}
Unless otherwise stated, results are reported from single runs. However, we used the six RC models discussed in Section~\ref{sec:rgf-filtering} to estimate cross-run variation. Using the procedure and code of \citet{sellam2021multiberts}, we find variation of about 0.5 points (F1). As such, we do not find differences smaller than this significant, and in our results focus only on larger effects.

\paragraph{Computational Budget and Environmental Impact}
We fine-tune all T5 models on Cloud TPU v3 hardware; each takes approximately 4 hours on 16 TPUs in pod configuration. 
Total compute time is approximately 96 TPU-hours and 192 GPU-hours, which we estimate as 43 kg CO2e using the method of \citet{luccioni2019quantifying}\footnote{\url{https://mlco2.github.io/impact/\#co2eq}}.

\section{Evaluation of Fluency and Noise}
\label{sec:appendix_human}
The authors sampled 300 examples of generated questions. To annotate for fluency, authors use the following rubric: Is the generated question grammatically well-formed barring non-standard spelling and capitalization of named entities. This noise annotation was done for RGF, as well as Gold Agen-Qgen and Random Agen-Qgen.

\paragraph{Creation of paired data for counterfactual evaluation}
Once again, authors annotate for correctness of counterfactual RGF instances that are paired by reference or predicate, as described in \S\ref{sec:filtering-eval}. Filtering is done until 1000 examples are available under each category.

\begin{table}[ht]
    \centering
    \begin{tabular}{c|c|c}
        \toprule
        \textbf{Data} & \textbf{Unfiltered} & \textbf{Filtered} \\ 
        \midrule
        RGF & 29.8\%	& 25.3\% \\
        Gold Agen-Qgen & 27.9\%	& 20.7\%  \\
        Random Agen-Qgen & 30.7\% & 28.3\% \\
        \bottomrule
    \end{tabular}
    \caption{Fraction of noise (incorrect $(q',c',a')$) in generated data, from 300 examples manually annotated by the authors.}
    \label{tab:noise_in_datasets}
\end{table}

\section{Additional Experiments}
\label{sec:additional_results}

\subsection{Intrinsic Evaluation}
\label{sec:additional_intrinsic}

\begin{figure}[ht!]
    \centering
    \includegraphics[width=\columnwidth]{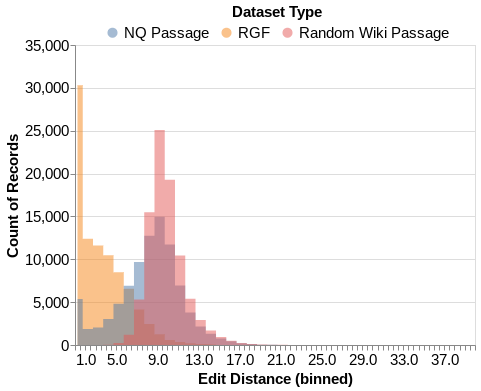}
    \caption{Distribution of edit distance between original $q$ and counterfactual $q'$ for RGF and other baselines for context selection. Note: For Random Wiki Passsage, original and generated questions bear no relation to each other and are randomly paired.}
    \label{fig:ed_distribution}
\end{figure}

In Figure~\ref{fig:ed_distribution}, we compare distributions of the edit distance between the original and generated questions for questions generated by our approach, those generated with the gold evidence passage (Gold Agen-Qgen baseline), and those generated from a random Wikipedia passage (\S\ref{sec:data_augmenation}) (Random Agen-Qgen baseline). We find that RGF counterfactuals undergo minimal perturbations from the original question compared to questions that are generated from random Wikipedia paragraph. 
This pattern also holds when compared to questions generated from gold NQ passages. We hypothesize that the set of alternate answers retrieved in our pipeline approach are semantically similar to the gold answer --- same entity type, for instance. 
Random answer spans chosen from the \emph{gold NQ passage} can result in significant semantic shifts in generated questions. 

\begin{figure}[ht!]
    \centering
    \includegraphics[width=\columnwidth]{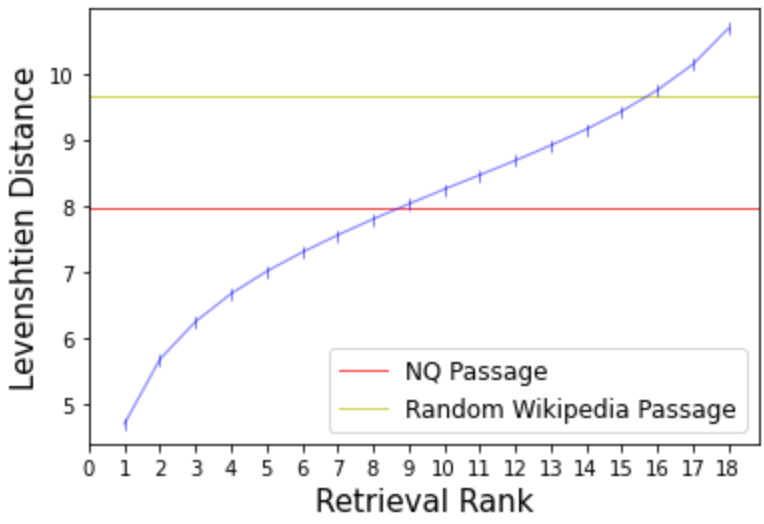}
    \caption{Plot of average edit distance between $q,q'$ vs. retrieval rank $r$, where $q'$ is generated from $r^{th}$ passage, showing that edit distance and retrieval rank are monotonically related.}
    \label{fig:edit_distance_rank}
\end{figure}

\begin{table*}[htb]
    \centering
    \begin{tabular}{c|l|c|c|c|c}
    \toprule
    \textbf{Consistency (OD)} & \textbf{Train Size} & \textbf{AQA}	& \textbf{AmbigQA} & \textbf{RGF $\Delta$ Ref.} & \textbf{RGF $\Delta$ Pred.} \\
    \midrule
    Original NQ &	90K	&	16.58	& 13.33	& 25.12 &	11.23 \\
    Random Agen-Qgen	& 90K + 90K &		15.80 &	20.00	& 27.94	 & 17.16 \\
    RGF (REALM-Qgen)	& 90K + 90K	 & 	\bf 17.66 & 	\bf 28.57	& \bf 31.77	& \bf 19.81 \\
    \bottomrule
    \end{tabular}
    \caption{Consistency Results for Open-domain QA.}
    \label{tab:consistency_odqa}
\end{table*}

\begin{table*}[htb]
    \centering
    \small
    \begin{tabular}{c|l|c|c|c|c|c|c|c}
    \toprule
    \textbf{Model} & \textbf{Train Size} & \textbf{NQ}	& \textbf{SQuAD} & \textbf{TriviaQA} & \textbf{HotpotQA} & \textbf{BioASQ} & \textbf{AQA} & \textbf{AmbigQA}\\
    \midrule
    RGF (1X) & 90K + 90K & 70.68 & 79.88 & 16.99 & 53.41  & \bf 44.88 & 28.20 & \bf 47.61 \\
    RGF (2X)	& 90K + 180K   &    \bf 70.78 &	\bf 80.33	& \bf 17.46 & \bf 54.09 &  44.75 & \bf 28.30 & 46.73 \\
    RGF (3X)	& 90K + 270K    & 	70.68 & 80.14	& 17.14 & 52.45 &  44.48 & 26.60 & 46.02 \\
    RGF (4X)	& 90K + 360K	 & 	70.17 & 79.97	& 17.06 & 51.82 &  44.35 & 27.50 & 46.81 \\
    \bottomrule
    \end{tabular}
    \caption{Reading comprehension results with varying training data augmentation sizes (exact match). We do not observe a consistent improvement with additional data. This series of experiments was run using an older version of T5X, so are not exactly comparable to Table~\ref{tab:rc_ood}.}
    \label{tab:varying_train_size}
\end{table*}

In Figure~\ref{fig:edit_distance_rank}, we measure the relation between retrieval rank and edit-distance for RGF. For retrieval rank $i$, we plot average edit distance between the original question and counterfactual question that was generated using the $i$th passage and answer. We observe a monotonic relation between retrieval rank and edit distance (which we use for filtering our training data). 
We also measure changes in the distribution of question type and predicate type between original NQ data and the generated RGF data.

\begin{figure}[ht!]
    \centering
    \includegraphics[width=\columnwidth]{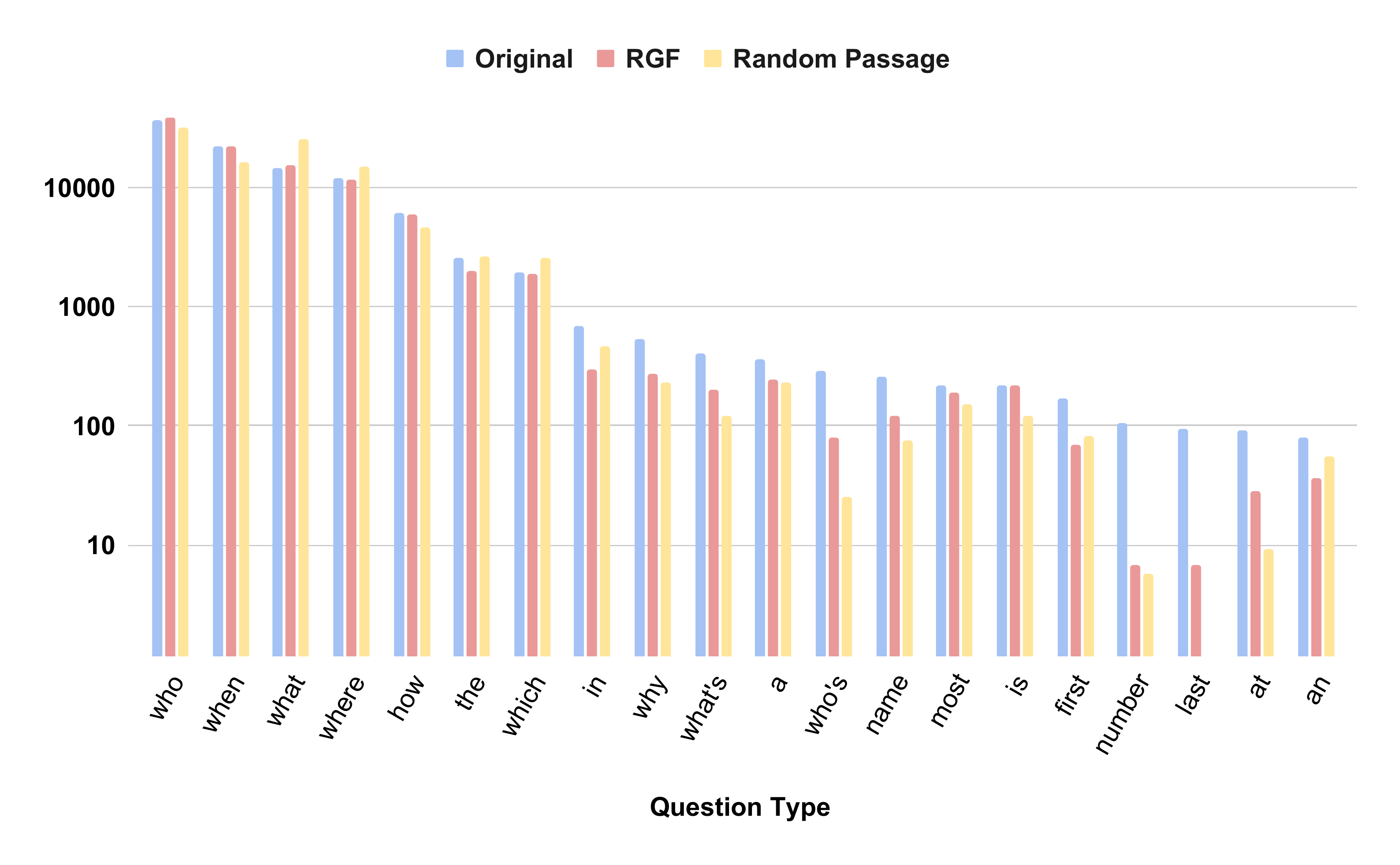}
    \caption{Distribution of top 20 question types for original NQ data, RGF counterfactuals and questions generated from random Wikipedia passage, indicating bias towards popular question types.}
    \label{fig:question_distribution}
\end{figure}

Figure~\ref{fig:question_distribution} indicates that counterfactual data exacerbates question-type bias. However, this bias exists in RGF as well as baselines.

\subsection{Consistency for Open-Domain QA}\label{results:odqa-consistency}
In Table~\ref{tab:consistency_odqa}, we show results on evaluating consistency on paired datasets in the open-domain results, similar to the results shown in \S\ref{results:robust} in the Reading Comprehension setting.

\subsection{Augmentation with more data}\label{results:rc-additional-training-data}
In Table~\ref{tab:varying_train_size}, we show results on augmenting with more than one RGF counterfactual triple $(q',a',c')$ for every original example $(q,a,c)$ in NQ. These experiments were run on an older version of T5X, so RGF (1X) values are reported differently from Table \ref{tab:rc_ood}. We observe that adding more RGF data (3X or more) for augmentation can hurt performance. This may be because of increase in the proportion of noisy to clean examples during training and exacerbation of biases in the question generation model (explored in \ref{fig:question_distribution}), resulting in diminishing returns. These challenges also occur in the baselines, and may be inhererent to augmentation with generated data.

\subsection{Effect of perturbation type}

\begin{table}[ht!]
    \centering
    \small
    \begin{tabular}{c|c|c|c}
        \toprule
        \bf Consistency (RC) 
        & \bf Val 1-4
        & \bf Val 5-10
        & \bf Val > 10 \\
        \midrule
        Train 1-4
        & 71.02
        & 67.55
        & 64.78 \\
        Train 5-10
        & 68.89
        & \bf 68.98
        & 63.92 \\
        Train >10
        & 65.78
        & 66.33
        & \bf 65.33 \\ 
        Train All 
        & \bf 72.34
        & 67.82
        & 65.12 \\
        \bottomrule
    \end{tabular}
    \caption{Results on sharding training data based on edit distance between $(q,q')$. Training dataset size for each bin is 90k NQ + 167k generated. Once again, training with all RGF data robustly improves consistency across different amounts of perturbations.}
    \label{tab:sharding_ed}
\end{table}

\paragraph{Experimental Setting}
For edit distance-based experiments, we shard training examples into three categories by binning word-level edit distance between $q$ and $q'$ into three ranges: $1$--$4$, $5$--$10$, and $>10$. We similarly categorize RGF data generated for the NQ development set into the same categories. Evaluation sets for edit-distance experiments based were not manually noise filtered. We again report consistency on the reading comprehension model.

\paragraph{Results} Similar to the observations for dataset sharding along QED annotations, when data is sharded by edit distance, we observe that using the full RGF data nearly matches the best performance from training on that shard, suggesting that CDA with the highly diverse RGF data can lead to improved consistency on a broad range of perturbation types.

\clearpage
\onecolumn
\section{Semantic Diversity}
\label{sec:appendix_semantic}
\begin{figure*}[h]
    \centering
    \includegraphics[scale=0.70]{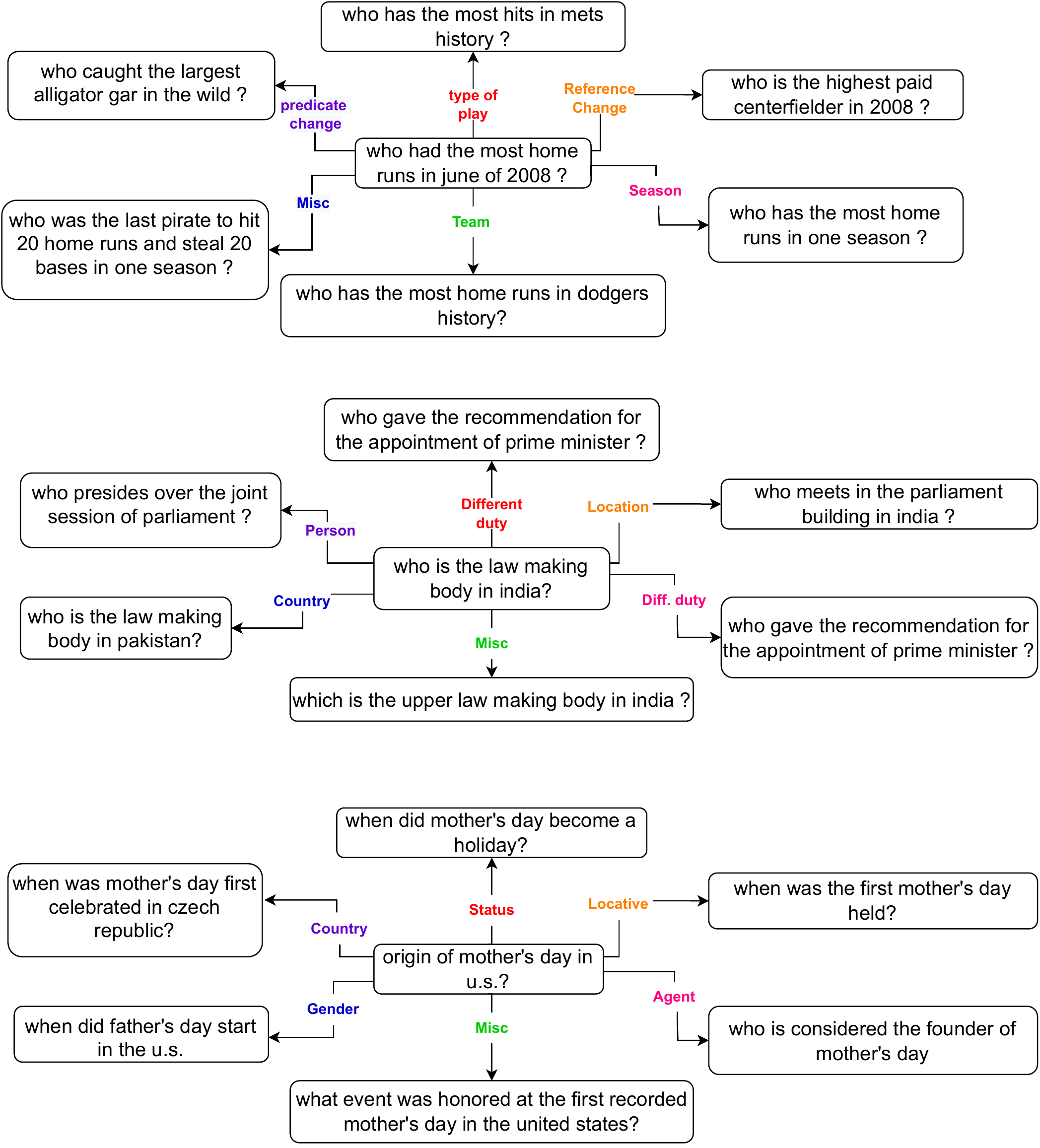}
    \caption{Context-specific semantic diversity of perturbations achieved by RGF on questions from NQ. The multiple latent semantic dimensions identified (arrows in the diagram) fall out of our retrieval-guided approach.}
    \label{fig:neighborhood_additional_examples}
\end{figure*}

Figure \ref{fig:neighborhood_additional_examples} includes more examples from Natural Questions, showing the counterfactual questions generated for different input questions by RGF.

\clearpage

\section{Error analysis of generated examples}
\label{sec:appendix_errors}
Table~\ref{tab:rgf_error_analysis} shows examples where the RGF model produced incorrect $(q', a', c')$ triples, selected from the manually-annotated subset described in Section~\ref{sec:intrinsic}.

\begin{table}[htb]
    \centering
    \begin{tabular}{p{0.9\linewidth}}
    \toprule
    \textbf{Nonsensical Question} \\
    \textbf{Context:} The security management process relates to other ITIL - processes . However , in this
    particular section the most obvious relations are the relations to the service level management , incident management and change management processes . Security management is a continuous process that can 
    be compared to \underline{W . Edwards Deming} ' s Quality Circle ( Plan , Do , Check , Act ) . The inputs are 
    requirements from clients . The requirements are translated into security services and security metrics.\\
    \textbf{Answer:} \underline{W . Edwards Deming}\\
    \textbf{Generated Question:} the security management process is similar to the itil ? \\
    \midrule
    \textbf{Incomplete Question} \\
    \textbf{Context:} Using Cartesian coordinates , inertial motion is described mathematically as : where " x " is the 
    position coordinate and " $\tau$ " is proper time . ( In Newtonian mechanics , " $\tau \equiv t$ " , the coordinate time ) . 
    In both Newtonian mechanics and special relativity , space and then spacetime are assumed to be flat ,  and we 
    can construct a global Cartesian coordinate system . In \underline{general relativity} , these restrictions on the shape of 
    spacetime and on the coordinate system to be used are lost . Therefore , a different definition of inertial  motion is required .\\
    \textbf{Answer:} \underline{general relativity} \\
    \textbf{Generated Question:} which theory states that all motion is a function of ? \\
    \midrule
    \textbf{Correct Type, but Wrong Entity}\\
    \textbf{Context:} Ruth McDevitt Ruth McDevitt ( September 13 , 1895  – May 27 , 1976 ) was an American stage , 
    film , radio and television actress . She was born Ruth Thane Shoecraft in Coldwater , Michigan . After attending the American Academy of Dramatic Arts , she married Patrick McDevitt and decided to devote
    her time to her marriage . After her husband ' s death in 1934 , she returned to acting . She performed on Broadway , in particular understudying and succeeding Josephine Hull in " Arsenic and Old Lace " and " The Solid Gold Cadillac " . She also worked as a radio actor . McDevitt was a familiar face on television during the 1950s , 1960s , and 1970s . She played " Mom Peepers " in the 1950s sitcom " Mister Peepers " . She was a regular with \underline{Ann Sheridan} , Douglas Fowley , and Gary Vinson in CBS ' s " Pistols ' n ' Petticoats " , a 1966 - 67 satire of the Old West .\\
    \textbf{Answer:} \underline{Ann Sheridan} \\
    \textbf{Generated Question:} who played the mother on mr peepers ? \\
    \bottomrule
    \end{tabular}
    \caption{Common error classes of RGF-generated $(q,c,a)$ triplets.}
    \label{tab:rgf_error_analysis}
\end{table}

\twocolumn
\clearpage

\end{document}